\documentclass[11pt,a4paper]{article}
\usepackage[hyperref]{acl2020}
\usepackage{times}
\usepackage{latexsym}
\usepackage{soul}
\usepackage[T1]{fontenc}
\usepackage[textwidth=0.7in,textsize=tiny]{todonotes}

\usepackage{microtype}
\usepackage{array}
\newcommand{\PreserveBackslash}[1]{\let\temp=\\#1\let\\=\temp}
\newcolumntype{C}[1]{>{\PreserveBackslash\centering}p{#1}}
\newcolumntype{R}[1]{>{\PreserveBackslash\raggedleft}p{#1}}
\newcolumntype{L}[1]{>{\PreserveBackslash\raggedright}p{#1}}

\aclfinalcopy % Uncomment this line for the final submission
\usepackage{amsmath}
\usepackage{amsfonts}
\usepackage{amssymb}
\usepackage{wrapfig}
\usepackage{subcaption}
\usepackage{multirow}
 \usepackage{mathtools} 

\usepackage{verbatim}

\usepackage{anyfontsize}

\usepackage{color}
\usepackage{tikz}
\usetikzlibrary{arrows,shapes,snakes,automata,backgrounds,fit,petri}
\usepackage{adjustbox}

\newcommand{\RN}[1]{%
	\textup{\lowercase\expandafter{\it \romannumeral#1}}%
}
\usepackage{soul}
\usepackage{tikz}
\usetikzlibrary{calc}
\usetikzlibrary{decorations.pathmorphing}

\makeatletter

\newcommand{\defhighlighter}[3][]{%
  \tikzset{every highlighter/.style={color=#2, fill opacity=#3, #1}}%
}

\defhighlighter{yellow}{.5}

\newcommand{\highlight@DoHighlight}{
  \fill [ decoration = {random steps, amplitude=1pt, segment length=15pt}
        , outer sep = -15pt, inner sep = 0pt, decorate
        , every highlighter, this highlighter ]
        ($(begin highlight)+(0,8pt)$) rectangle ($(end highlight)+(0,-3pt)$) ;
}

\newcommand{\highlight@BeginHighlight}{
  \coordinate (begin highlight) at (0,0) ;
}

\newcommand{\highlight@EndHighlight}{
  \coordinate (end highlight) at (0,0) ;
}

\newdimen\highlight@previous
\newdimen\highlight@current

\DeclareRobustCommand*\highlight[1][]{%
  \tikzset{this highlighter/.style={#1}}%
  \SOUL@setup
  \def\SOUL@preamble{%
    \begin{tikzpicture}[overlay, remember picture]
      \highlight@BeginHighlight
      \highlight@EndHighlight
    \end{tikzpicture}%
  }%
  \def\SOUL@postamble{%
    \begin{tikzpicture}[overlay, remember picture]
      \highlight@EndHighlight
      \highlight@DoHighlight
    \end{tikzpicture}%
  }%
  \def\SOUL@everyhyphen{%
    \discretionary{%
      \SOUL@setkern\SOUL@hyphkern
      \SOUL@sethyphenchar
      \tikz[overlay, remember picture] \highlight@EndHighlight ;%
    }{%
    }{%
      \SOUL@setkern\SOUL@charkern
    }%
  }%
  \def\SOUL@everyexhyphen##1{%
    \SOUL@setkern\SOUL@hyphkern
    \hbox{##1}%
    \discretionary{%
      \tikz[overlay, remember picture] \highlight@EndHighlight ;%
    }{%
    }{%
      \SOUL@setkern\SOUL@charkern
    }%
  }%
  \def\SOUL@everysyllable{%
    \begin{tikzpicture}[overlay, remember picture]
      \path let \p0 = (begin highlight), \p1 = (0,0) in \pgfextra
        \global\highlight@previous=\y0
        \global\highlight@current =\y1
      \endpgfextra (0,0) ;
      \ifdim\highlight@current < \highlight@previous
        \highlight@DoHighlight
        \highlight@BeginHighlight
      \fi
    \end{tikzpicture}%
    \the\SOUL@syllable
    \tikz[overlay, remember picture] \highlight@EndHighlight ;%
  }%
  \SOUL@
}
\makeatother
\newcommand{\hlightPyellow}[1]{%
	\ooalign{\hss\makebox[1pt]{\fcolorbox{orange!40}{yellow!40}{$#1$}}\hss\cr\phantom{$#1$}}%
}
\newcommand{\hlightPred}[1]{%
	\ooalign{\hss\makebox[1pt]{\fcolorbox{red!100}{red!30}{$#1$}}\hss\cr\phantom{$#1$}}%
}
\newcommand{\hlightPblue}[1]{%
	\ooalign{\hss\makebox[1pt]{\fcolorbox{blue!100}{blue!30}{$#1$}}\hss\cr\phantom{$#1$}}%
}
\newcommand{\hlightPgreen}[1]{%
	\ooalign{\hss\makebox[1pt]{\fcolorbox{green!100}{green!20}{$#1$}}\hss\cr\phantom{$#1$}}%
}

\newcommand*\circled[1]{\tikz[baseline=(char.base)]{
            \node[shape=circle,draw,inner sep=0.6pt] (char) {#1};}}
            
\usepackage{booktabs}

\usepackage{tcolorbox}

\makeatletter
\newcommand{\distas}[1]{\mathbin{\overset{#1}{\kern\z@\sim}}}%

\usepackage{enumitem}

\usepackage[lined,boxed,commentsnumbered,ruled,linesnumbered]{algorithm2e}

\newcommand{\ie}[0]{\emph{i.e., }}

\newcommand{\eg}[0]{\emph{e.g., }}

\newcommand{\beq}{\vspace{0mm}\begin{equation}}
\newcommand{\eeq}{\vspace{0mm}\end{equation}}
\newcommand{\beqs}{\vspace{0mm}\begin{eqnarray}}
\newcommand{\eeqs}{\vspace{0mm}\end{eqnarray}}
\newcommand{\barr}{\begin{array}}
\newcommand{\earr}{\end{array}}

\newcommand{\Wmat}[0]{{{\bf W}}}
\newcommand{\Xmat}[0]{{{\bf X}}}
\newcommand{\Ymat}{{\bf Y}}

\newcommand{\bv}[0]{{\boldsymbol{b}}}
\newcommand{\cv}[0]{{\boldsymbol{c}}}

\newcommand{\xv}{\boldsymbol{x}}
\newcommand{\yv}{\boldsymbol{y}}
\newcommand{\zv}{\boldsymbol{z}}

\newcommand{\thetav}{\boldsymbol{\theta}}

\newcommand{\E}{\mathbb{E}}

\newcommand{\I}{\mathbb{I}}

\usepackage[normalem]{ulem}

\newcommand{\Ycal}{\mathcal{Y}}
\newcommand{\Lcal}{\mathcal{L}}

\newcommand{\Dcal}{\mathcal{D}}

\newcommand{\Qcal}{\mathcal{Q}}

\newcommand{\Scal}{\mathcal{S}}

\ifx\assumption\undefined
\newtheorem{assumption}{Assumption}
\fi

\ifx\definition\undefined
\newtheorem{definition}{Definition}
\fi

\ifx\remark\undefined
\newtheorem{remark}{Remark}
\fi

\usepackage{color, colortbl}
\definecolor{Gray}{gray}{0.93}

\DeclareMathOperator*{\argmin}{arg\,min}

\title{Few-Shot Named Entity Recognition: A Comprehensive Study}

\author{
Jiaxin Huang$^{1}\!$
\thanks{\hspace{1mm} Work performed during an internship at Microsoft~~ \hspace{1mm}}~,
 Chunyuan Li$^{2}\!$
\thanks{\hspace{1mm}  Corresponding author~~ \hspace{1mm}}~, 
~Krishan Subudhi$^{2}$,
~Damien Jose$^{2}$,
\\
\bf{
Shobana Balakrishnan$^{2}$,
~Weizhu Chen$^{2}$,
~Baolin Peng$^{2}$, ~Jianfeng Gao$^{2}$, ~Jiawei Han$^{1}$}
\\ $^{1}$University of Illinois Urbana-Champaign ~~~ $^{2}$Microsoft \\
{\tt\small \{jiaxinh3, hanj\}@illinois.edu ~~~ }\\
{\tt\small \{chunyl,krkusuk,dajose,shobanab,wzchen,bapeng,jfgao\}@microsoft.com}
}

\date{}

\begin{document}
\maketitle

\setlength{\abovedisplayskip}{4pt}
\setlength{\belowdisplayskip}{4pt}

\begin{abstract}
This paper presents a comprehensive study to efficiently build named entity recognition (NER) systems when a small number of in-domain labeled data is available.
Based upon recent Transformer-based self-supervised pre-trained language models (PLMs), we investigate three orthogonal schemes to improve the model generalization ability for few-shot settings: (1) meta-learning to construct prototypes for different entity types, (2) supervised pre-training on noisy web data to extract entity-related generic representations and (3) self-training to leverage unlabeled in-domain data. Different combinations of these schemes are also considered. 
We perform extensive empirical comparisons on 10 public NER datasets with various proportions of labeled data, suggesting useful insights for future research. Our experiments show that 
$(\RN{1})$ in the few-shot learning setting, 
the proposed NER schemes significantly improve or
outperform the commonly used baseline, a PLM-based linear classifier fine-tuned on domain labels.
$(\RN{2})$ We create new state-of-the-art results on both few-shot and training-free settings compared with existing methods. 
We will release our code and pre-trained models for reproducible research.
\end{abstract}

\vspace{0mm}
\section{Introduction}
Named Entity Recognition (NER) involves processing unstructured text, locating and classifying named entities (certain occurrences of words or expressions) into particular categories of pre-defined entity types, such as persons, organizations, locations, medical codes, dates and quantities.
NER serves as an important first component for tasks such as information extraction~\cite{Ritter2012OpenDE}, information retrieval~\cite{Guo2009NamedER}, question answering~\cite{molla2006named}, task-oriented dialogues~\cite{peng2020soloist,gao2019neural} and other language understanding applications~\cite{nadeau2007survey,shaalan2014survey}.
Deep learning has shown remarkable success in NER in recent years, especially with self-supervised pre-trained language models (PLMs) such as BERT~\cite{devlin2019bert} and RoBERTa~\cite{liu2019roberta}. State-of-the-art (SoTA) NER models are often initialized with PLM weights, fine-tuned with standard supervised learning. One classical approach is to add a linear classifier on top of representations provided by PLMs, and fine-tune the entire model with a cross-entropy objective on domain labels~\cite{devlin2019bert}. Desipite its simplicity, the approach provides strong results on several benchmarks and is served as baseline in this study.  
% Though simple and effective, 

% More complicated methods further add Conditional Random Fields (CRF) to encourage structural learning between tokens. 

Unfortunately, even with these PLMs, building NER systems remains a labor-intensive, time-consuming task. It requires rich domain knowledge and expert experience to annotate a large corpus of in-domain labeled tokens to make the models work well.  However, this is in contrast to the real-world application scenarios, where only very limited amounts of labeled data are available for new domains. For example, a new customer would prefer to provide very few labeled examples for specific domains in cloud-based NER services.
The cost of building NER systems at scale with rich annotations (\ie hundreds of different enterprise use-cases/domains) can be prohibitively expensive. This draws attentions to a challenging but practical research problem: few-shot NER.

\begin{figure}[t!]
\vspace{2mm}
\centering
{\includegraphics[width=0.38\textwidth]{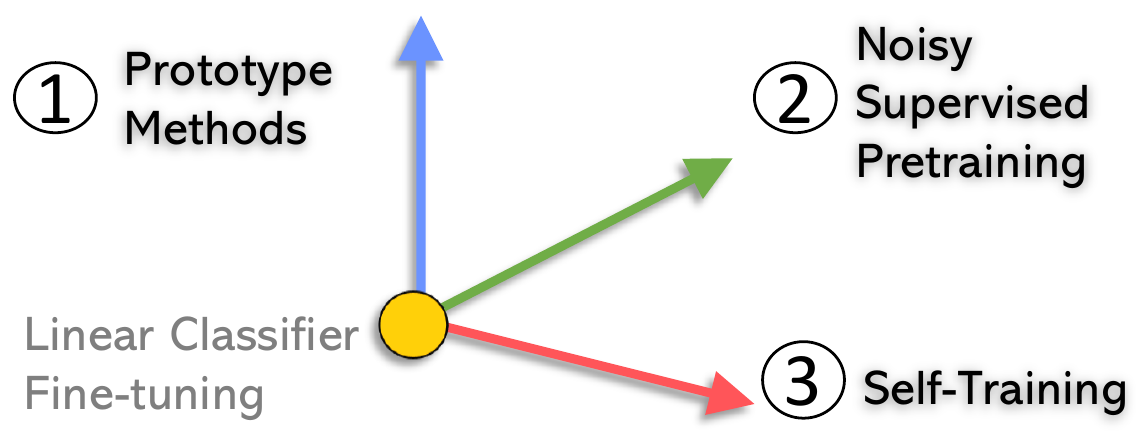}}
\vspace{-0mm}
\caption{An overview of methods studied in our paper. Linear classifier fine-tuning is a default baseline that updates an NER model from pre-trained Roberta/BERT. We study three orthogonal strategies to improve NER models in the limited labeled data settings.}
\label{fig:fewshotNER_scheme}
\vspace{-2mm}
\end{figure}

To deal with the challenge of few-shot learning, we focus on improving the generalization ability of PLMs for NER from three complementary directions, shown in Figure~\ref{fig:fewshotNER_scheme}. Instead of limiting ourselves in making use of limited in-domain labeled tokens with the classical approach,  
$(\RN{1})$ we create prototypes as the representations for different entity types, and assign labels via the nearest neighbor criterion;
$(\RN{2})$ we continuously pre-train PLMs using web data with noisy labels that is available in much larger quantities to improve NER accuracy and robustness; 
$(\RN{3})$ we employ unlabeled in-domain tokens to predict their soft labels using self-training, and perform semi-supervised learning in conjunction with the limited labeled data.

Our contributions include:
$(\RN{1})$ We present the first systematic study for few-shot NER, a problem that is previously little explored in the literature.  Three distinctive schemes and their combinations are investigated.  
$(\RN{2})$ We perform comprehensive comparisons of these schemes on 10 public NER datasets from different domains. 
$(\RN{3})$ Compared with existing methods on few-shot and training-free NER settings , the proposed schemes achieve SoTA performance despite their simplicity.
To shed light on future research on few-shot NER, our study suggests that: 
$(\RN{1})$ Noisy supervised pre-training can significantly improve NER accuracy, and we will release our pre-trained checkpoints.
$(\RN{2})$ Self-training consistently improves few-shot learning when the ratio of data amounts between unlabeled and labeled data is high.
$(\RN{3})$ The performance of prototype learning varies on different datasets. It is useful when the number of labeled examples is small, or when new entity types are given in the training-free settings.

\section{Background on Few-shot NER}
\paragraph{Few-shot NER.} 
 a sequence labeling task, where the input is a text sequence (\eg sentence) of length $T$, $\Xmat=[\xv_1, \xv_2, ..., \xv_T ]$, and the output is a corresponding $T$-length labeling sequence $\Ymat=[\yv_1, \yv_2, ..., \yv_T]$, $ \yv  \in \Ycal$ is a one-hot vector indicating the entity type of each token from a pre-defined discrete label space. The training dataset for NER often consists of pair-wise data $\Dcal^{\mathtt{L}} = \{ (\Xmat_n, \Ymat_n) \}_{n=1}^N $, where $N$ is the number of training examples. Traditional NER systems are trained in the standard supervised learning paradigms, which usually requires a large number of pairwise examples, \ie $N$ is large.
In real-world applications, the more favorable scenarios are that only a small number of labeled examples are given for each entity type ($N$ is small), because expanding labeled data increases annotation cost and decreases customer engagement. This yields a challenging task {\it few-shot NER}.
%  Specifically, in $k$-shot learning setting on a target domain corpus $\Dcal_t$ with entity type set $\Ycal_t$, the support set $\Scal_t$ consists of $k$ support examples per entity type. The task is challenging due to the scarcity of labeled data, thus few-shot NER systems are typically first pre-trained on one or more source domains $\Dcal_s$ with source entity type set $\Ycal_s$ to learn to capture type-related information and generalize to target domains. \CL{I am NOT sure it is a good idea to introduce the notion of support/query set here. I suggest we introduce them in the section of prototype method}

\begin{figure*}[t!]%\vspace{-25pt}
	\vspace{-0mm}\centering
	\begin{tabular}{c c}
		\hspace{-4mm}
		\includegraphics[height=2.4cm,width=7.20cm]{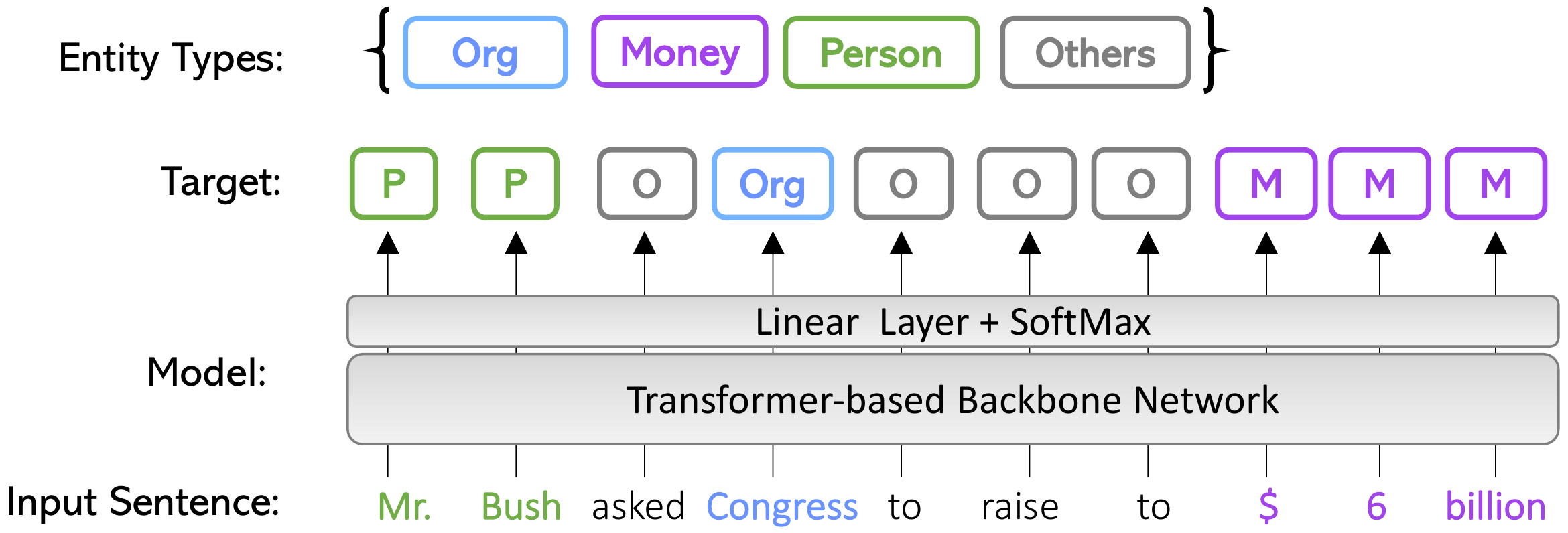} & 
		% \vspace{2mm}
		\hspace{-3mm}
		\includegraphics[height=2.1cm,width=7.70cm]{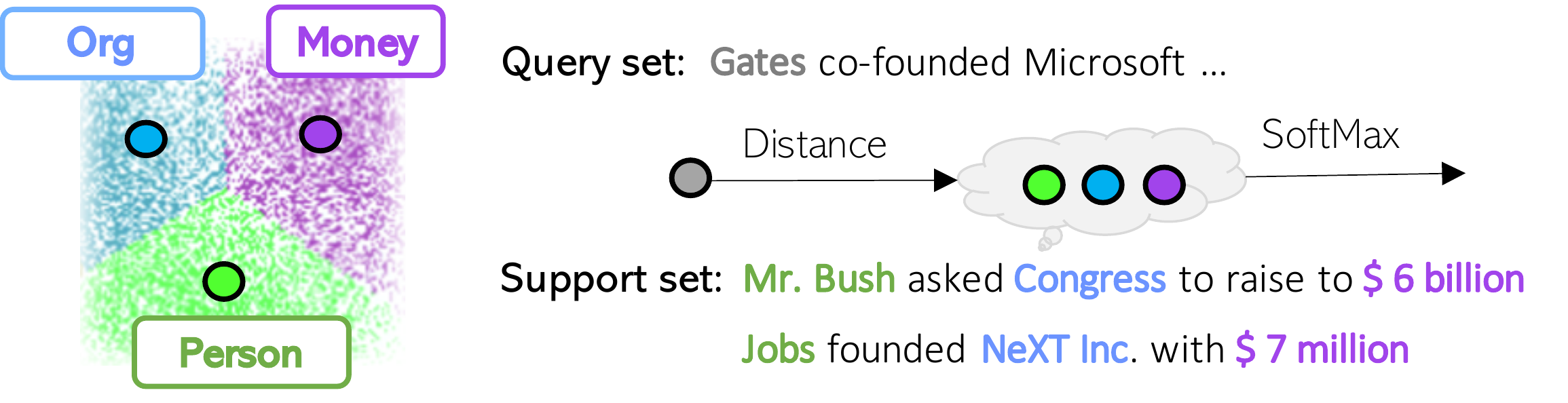} \\
		\vspace{0mm}
		\hspace{-6mm}
		(a) Baseline: NER with a linear classifier \hspace{3mm} & 
		(b) Prototype-based method   \vspace{2mm}\\ 
		\hspace{-4mm}
		\includegraphics[height=2.1cm,width=7.60cm]{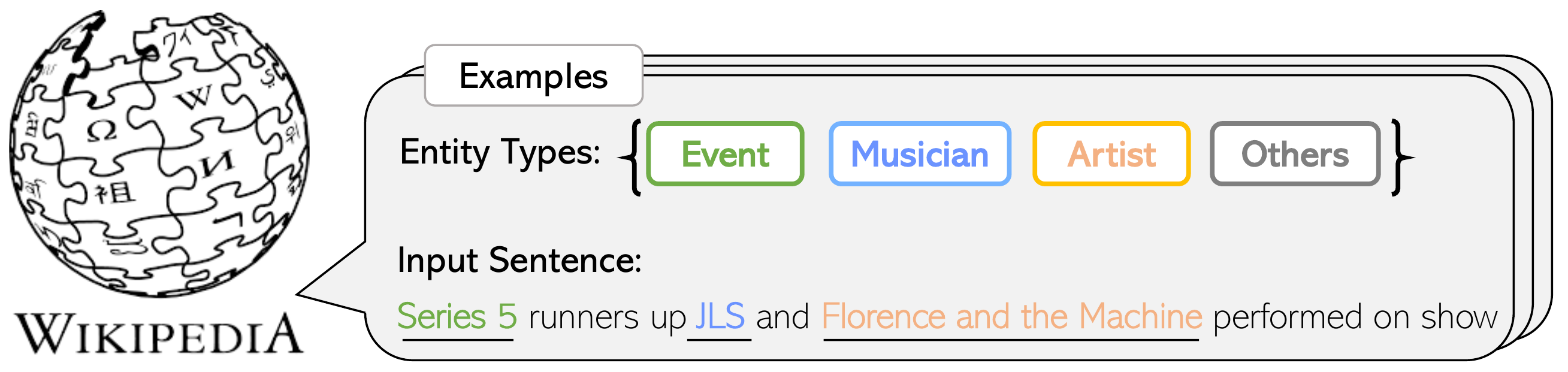} & 
		\hspace{3mm}
		\includegraphics[height=2.7cm,width=7.00cm]{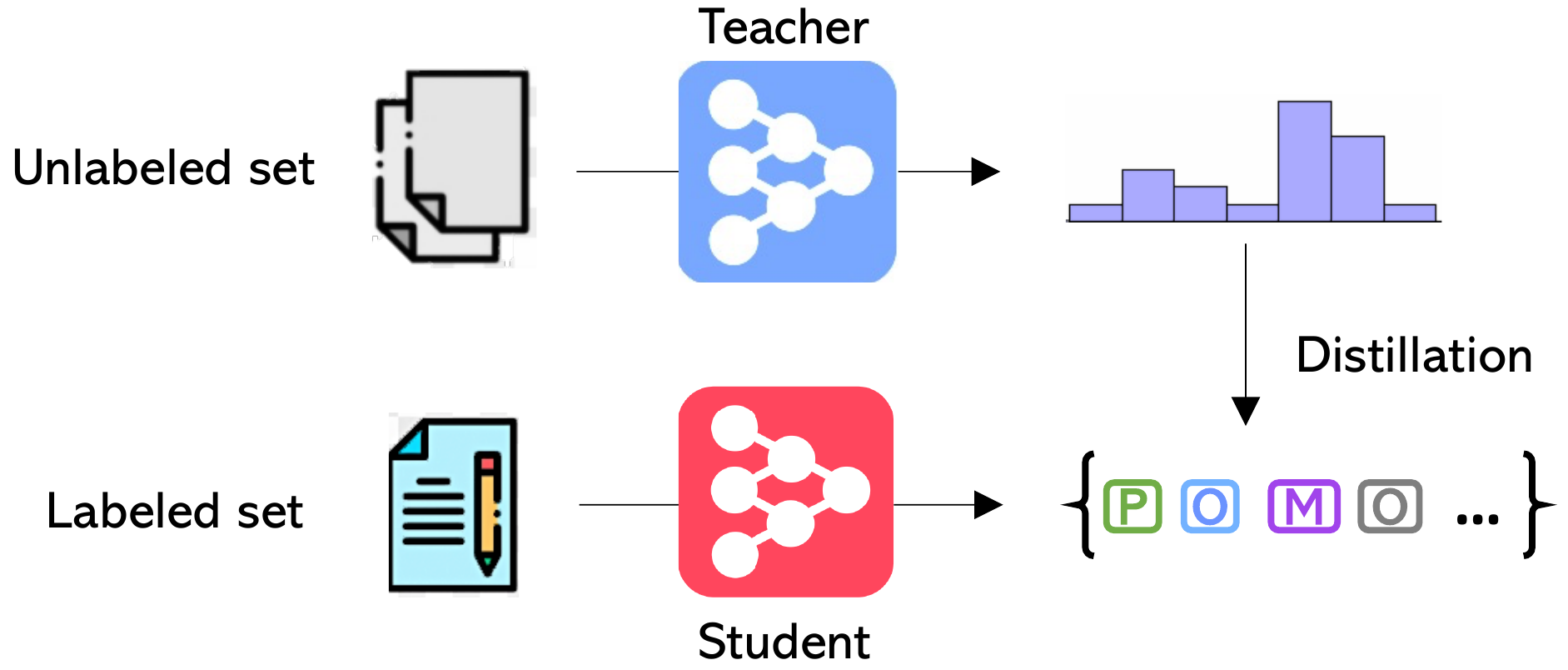} \\
		(c) Noisy supervised pre-training  \vspace{0mm} & 
		(d) Self-training  \hspace{-0mm} \\ 
	\end{tabular}
	\vspace{-0mm}
	\caption{Illustration of different methods for few-shot NER. In this example, each token in the input sentence is categorized into one of the four entity types. (a) A typical NER system, where a linear classifier is built on top of unsupervised pre-trained Transformer-based networks such as BERT/Roberta. (b) A prototype set is constructed via averaging features of all tokens belonging to a given entity type in the support set (\eg the prototype for \texttt{Person} is an average of three tokens: \textit{Mr.}, \textit{Bush} and \textit{Jobs}). For a token in the query set, its distances from different prototypes are computed, and the model is trained to maximize the likelihood to assign the query token to its target prototype. (c) The Wikipedia dataset is employed for supervised pre-training, whose entity types are related but different (\eg \texttt{Musician} and \texttt{Artist} are more fine-grained types of \texttt{Person} in the downstream task). The associated types on each token can be noisy. (d) Self-training: An NER system (teacher model) trained on a small labeled dataset is used to predict soft labels for sentences in a large unlabeled dataset. The joint of the predicted dataset and original dataset is used to train a student model. 
	 }
	\vspace{-2mm}
	\label{fig:ner_schemes}
\end{figure*}

\paragraph{Linear Classifier Fine-tuning.}
Following the recent self-supervised PLMs~\cite{devlin2019bert,liu2019roberta}, a typical method for NER is to utilize a Transformer-based backbone network to extract the contextualized representation of each token $ \zv = f_{\thetav_0}(\xv)$ . A linear classifier (\ie a linear layer with parameter $\thetav_1 = \{\Wmat, \bv\} $  followed by a Softmax layer) is applied to project the representation $\zv$ into the label space $f_{\thetav_1}(\zv) = \text{Softmax}(\Wmat \zv + \bv)$. In another word, the end-to-end learning objective for linear classifier based NER can be obtained via a function composition  $ \yv = f_{\thetav_1} \circ f_{\thetav_0}(\xv)$, with trainable parameters $\thetav = \{ \thetav_0, \thetav_1 \}$.  The pipeline is shown in Figure \ref{fig:ner_schemes}(a). 
The model is optimized by minimizing the cross-entropy:
\begin{equation}\label{eq:mle}
    \mathcal{L}(\xv, \yv) = \sum_{ (\Xmat, \Ymat) \in \Dcal^{\mathtt{L}}} \sum_{i=1}^T \text{KL}( \yv_i || q(\yv_i |\xv_i) ), 
\end{equation}
where the KL divergence between two distribution is $\text{KL}(p || q)= \E_p \log (p/q) $, and the prediction probability vector for each token is
\begin{equation}\label{eq:naiveft}
    q(\yv | \xv)=\text{Softmax}(\Wmat\cdot f_{\thetav_0}(\xv) + \bv)
\end{equation}

In practice,  $\thetav_1 = \{\Wmat, \bv\} $  is always updated, 
% \jg{$\thetav_1$ does not occur in Equation 2.}, 
while $\thetav_0$ can be either frozen~\cite{Liu2019TowardsIN,Liu2019GCDTAG,Jie2019DependencyGuidedLF} or updated~\cite{devlin2019bert,Yang2020SimpleAE}. 

% More complicated methods such as Conditional Random Fields (CRF) have been studied to encourage structural learning between tokens. We empirically compare with it in the experiments.

% few-shot NER 

% SoftMax Fine-tuning

% However, though it is simple and effective, 

\section{Methods}

When only a small number of labeled tokens are available, it renders difficulties for the 
%traditional Softmax 
classical supervised
fine-tuning approach: the model tends to over-fit the training examples and shows poor generalization performance on the testing set~\cite{Fritzler2019FewshotCI}. 
% \jg{better to provide references.}
In this paper, we provide a comprehensive study specifically for limited NER data settings, and explore three orthogonal directions shown in Figure \ref{fig:fewshotNER_scheme}: 
$(\RN{1})$ How to adapt meta-learning such as prototype-based methods for few-shot NER?
$(\RN{2})$ How to leverage freely-available web data as noisy supervised pre-training data?
$(\RN{3})$ How to leverage unlabeled in-domain sentences in a semi-supervised manner?
 Note that these three directions are complementary to each other, can be further used jointly to extrapolate the methodology space in Figure \ref{fig:fewshotNER_scheme}.

\vspace{-1mm}
\subsection{Prototype-based Methods}
\vspace{-1mm}

% (introduce episode learning)
Meta-learning~\cite{Ravi2017OptimizationAA} have shown promising results for few-shot image classification~\cite{Tian2020RethinkingFI} and sentence classification~\cite{Yu2018DiverseFT,Geng2019InductionNF}. It is natural to adapt this idea to few-shot NER. The core idea is to use episodic classification paradigm to simulate few-shot settings during model training. Specifically in each episode, $M$ entity types (usually $M<|\Ycal|$) are randomly sampled from $\Dcal^{\mathtt{L}}$, containing a \textit{support} set $\mathcal{S}=\{(\Xmat_i, \Ymat_i \}_{i=1}^{M\times K}$ ($K$ sentences per type) and a \textit{query} set $\mathcal{Q}=\{(\hat{\Xmat}_i, \hat{\Ymat}_i \}_{i=1}^{M\times K'}$ ($K'$ sentences per type). 

We build our method based on prototypical network~\cite{Snell2017PrototypicalNF}, which introduces the notion of {\it prototypes}, representing entity types as vectors in the same representation space of individual tokens.
To construct the prototype for the $m$-th entity type $\cv_m$, the average of representations is computed for all tokens belonging to this type in the support set $\Scal$:
% \CL{bad use of $k$, it has been used Section 2}
\begin{equation}
    \cv_m = \frac{1}{|\Scal_m|}\sum_{\xv\in \Scal_m} f_{\thetav_0}(\xv),
\end{equation}
where $\Scal_m$ is the tokens set of the $m$-th type in $\Scal$, and $f_{\thetav_0}$ is defined in \eqref{eq:naiveft}.
For an input token $\xv \in \Qcal$ from the query set, its prediction distribution is computed by a softmax function of the distance between $\xv$ and all the entity prototypes. For example, the prediction probability for the $m$-th prototype is:
\begin{equation}
~\hspace{-1mm}
    q(\yv\!=\!\I_m|\xv)\!=\!\frac{\exp{(-d(f_{\thetav_0}(\xv), \cv_m))}}{\sum_{m'}\exp{(-d(f_{\thetav_0}(\xv),\cv_{m'}))}}
    \label{eq:prototype_knn}
\end{equation}
where $\I_m$ is the one-hot vector with 1 for $m$-th coordinate and 0 elsewhere, and $d(f_{\thetav_0}(\xv),\cv_{m}) = \|f_{\thetav_0}(\xv) - \cv_{m}  \|_2 $ is used in our implementation.
We provide a simple example to illustrate the prototype method in Figure~\ref{fig:ner_schemes}(b).
In each training iteration, a new episode is sampled, and  the model parameter $\thetav_0$ is updated via plugging \eqref{eq:prototype_knn} into \eqref{eq:mle}. In testing phase, the label of a new token $\xv$ is assigned using the nearest neighbor criterion $\argmin_m d(f_{\thetav_0}(\xv),\cv_{m})$.

\vspace{-1mm}
\subsection{Noisy Supervised Pre-training}\label{sec:nsp}
\vspace{-1mm}
Generic representations via self-supervised pre-trained language models~\cite{devlin2019bert,liu2019roberta} have benefited a wide range of NLP applications. These models are pre-trained with the task of randomly masked token prediction on massive corpora, and are agnostic to the downstream tasks. In other words, PLMs treat each token equally, which is not aligned with the goal of NER: identifying named entities as emphasized tokens and assigning labels to them. For example, for a sentence ``$\textit{ Mr. Bush asked Congress to raise to $\$$ 6 billion }$'', PLMs treat \textit{to} and \textit{Congress} equally, while NER aims to highlight entities like \textit{Congress} and lowlight their collocated non-entity words like \textit{to}.

% and typically people start at such pre-trained language models to directly fine-tune the models on downstream tasks such as named entity recognition or text classification. Though demonstrating superior performance, in the few-shot learning setting where there are limited training data in the target domain, directly applying the fine-tuning approach may not guarantee the best performance since the contextualized representation is sub-optimal for target tasks like named entity recognition. 

This intuition inspires us to endow the backbone network an ability to  outweigh the representations of entities for NER. Hence, we propose to employ the large-scale noisy web data $\mathtt{WiNER}$ \cite{Ghaddar2017WiNERAW} for noisy supervised pre-training (NSP). The labels in $\mathtt{WiNER}$ are automatically annotated on the 2013 English Wikipedia dump by querying anchored strings as well as their coreference mentions in each wiki page to the Freebase. The $\mathtt{WiNER}$ dataset is of 6.8GB and contains 113 entity types along with over 50 million sentences. Though introducing inevitable noises (\eg a random subset of 1000 mentions are manually evaluated and the accuracy of automatic annotations reaches 77\%, due to the error of identifying coreferences), this automatic annotation procedure is highly scalable and affordable.
The label set of $\mathtt{WiNER}$ covers a wide range of entity types. They are often related but different from entity types in the downstream datasets. For example in Figure~\ref{fig:ner_schemes}(c), the entity types \texttt{Musician} and \texttt{Artist} in Wikipedia are more fine-grained than \texttt{Person} in a typical NER dataset. The proposed NSP learns representations to distinguish entities from others. This particularly favors the few-shot settings, preventing over-fitting via the prior knowledge of extracting entities from various contexts in pre-training.

% Previous works in machine learning have already shown the effectiveness of large-scale supervised pre-training on object detection and blabla\cite{}. Here we propose to add an intermediate stage called noisy supervised pre-training between the self-supervised pre-training and downstream task fine-tuning to learn a task-related text encoder. \CL{Go to related work}

% Though the large pre-trained dataset includes inevitable noise, we later show that models pre-trained on it improve the performance across all the target-domain datasets.

Two pre-training objectives are considered in NSP, respectively: the first one is to use the linear classifier in \eqref{eq:naiveft}, the other is a prototype-based objective in~\eqref{eq:prototype_knn}. For the linear classifier, we found that the batch size of $1024$ and learning rate of $1e^{-4}$ works best, and for the prototype-based approach, we use the episodic training paradigm with $M=5$ and set learning rate to be $5e^{-5}$. For both objectives, we train the whole corpus for $1$ epoch and apply the Adam Optimizer~\cite{Kingma2015AdamAM} with a linearly decaying schedule with warmup at $0.1$.
We empirically compare both objectives in experiments, and found that the linear classifier in \eqref{eq:naiveft} improves pre-training more significantly. 
% \jg{To clarify: NSP is a \emph{continual} pre-training method. So we might say that NSP improves the pre-trained model.}
% \CL{Please provide WiNER pre-training details here, \#token and \#labels, pre-training schedule}
% \CL{I am considering to introduce prototype-based methods first so that we have the equations to refer}. 

\vspace{-1mm}
\subsection{Self-training}
Though manually labeling entities is expensive, it is easy to collect large amounts of unlabeled data in the target domain. Hence, it becomes desired to improve the model performance by effectively leveraging unlabeled data $\Dcal^{\mathtt{U}}$ with limited labeled data $\Dcal^{\mathtt{L}}$. We resort to the recent self-training scheme~\cite{Xie2020SelfTrainingWN} for semi-supervised learning. % student-teacher network that consists of a student model and a teacher model to iteratively enhance each other. 
The algorithm operates as follows:

\begin{enumerate}
\item
Learn teacher model $\thetav^{\mathtt{tea}}$ via cross-entropy using \eqref{eq:mle} with labeled tokens $\Dcal^{\mathtt{L}}$.

\item
Generate soft labels using a teacher model on unlabeled
tokens:
\begin{equation}\label{eq:soft_labels}
    \tilde{\yv}_i=f_{\thetav^{\mathtt{tea}}} (\tilde{\xv}_i ), \forall \tilde{\xv}_i \in  \Dcal^{\mathtt{U}}
\end{equation}

\item
Learn a student model $\thetav^{\mathtt{stu}}$ via cross-entropy using \eqref{eq:mle} on labeled and unlabeled tokens:
\begin{align}
\Lcal_{\text{ST}} = 
   &  \frac{1}{|\Dcal^{\mathtt{L}}|}  \sum_{\xv_i \in\Dcal^{\mathtt{L}}} \Lcal( f_{\thetav^{\mathtt{stu}}} (\xv_i), \yv_i ) \nonumber\\
   & + \frac{\lambda_{\mathtt{U}}}{|\Dcal^{\mathtt{U}}|} \sum_{ \tilde{\xv}_i \in \Dcal^{\mathtt{U}} } \Lcal( f_{\thetav^{\mathtt{stu}}} (\tilde{\xv}_i),  \tilde{\yv}_i)
\end{align}

where $\lambda_{\mathtt{U}}$ is the weighting hyper-parameter.
\end{enumerate}
A visual illustration for self-training procedure shown in Figure~\ref{fig:ner_schemes}(d).
It is optional to iterate from Step 1 to  Step 3 multiple times, by initializing $\thetav^{\mathtt{tea}}$ in Step 1 with newly learned $\thetav^{\mathtt{stu}}$ in Step 3. We only perform self-training once in our experiments for simplicity, which has already shown excellent performance.  
% We use $g(\cdot;\theta_{\text{tea}})$ to denote the teacher model and $g(\cdot;\theta_{\text{stu}})$ to denote the student model.
% Starting from a trained classification model $g(\cdot;\theta_0)$ on limited supervised data, we initialize the teacher model as 
% \begin{equation}
%     \theta_{\text{tea}}^{0}=\theta_0
% \end{equation}
% At iteration $t$, we generate soft labels for each unlabeled data $\tilde{x}_i$ by

% where $s_i$ is the $i$-th sentence in the unlabeled dataset.
% Then the student model is trained to fit both the hard labels of limited supervised data and the soft labels of unlabeled data.
% At the end of iteration, $\theta_{\text{tea}}^{t+1}$ is set to $\theta_{\text{stu}}^{t}$ to be a new teacher in the next iteration.

\section{Related Work}

\paragraph{General NER.} NER is a long standing problem in NLP. 
Deep learning has significantly improve the recognition accuracy. Early efforts include exploring various neural architectures~\cite{lample2016neural} such as Bidrectional LSTMs~\cite{chiu2016named} and adding CRFs to capture structures~\cite{ma2016end}. Early studies have noticed the importance of reducing the annotation labor, where semi-supervised learning is employed, such as clustering~\cite{lin2009phrase}, and combining supervised objective with unsupervised word representations~\cite{turian2010word}. PLMs have recently revolutionized NER, where large-scale Transformer-based architectures~\cite{Peters2018DeepCW,devlin2019bert} are used as backbone network to extract informative representations.
Contextualized string embedding~\cite{Akbik2018ContextualSE} is proposed to capture subword structures and polysemous words in different usage.
Masked words and entities are jointly trained for prediction in~\cite{Yamada2020LUKEDC} with entity-aware self-attention.
These methods are designed for standard supervised learning, and have a limited generalization ability in few-shot settings, as empirically shown in~\cite{Fritzler2019FewshotCI}.

\paragraph{Prototype-based methods} recently become popular few-shot learning approaches in machine learning community.  It was firstly studied in the context of image classification~\cite{vinyals2016matching,sung2018learning,zhao2020remp}, and has recently been adapted to different NLP tasks such as text classification~\cite{wang2018joint,Geng2019InductionNF,bansal2020self}, machine translation~\cite{Gu2018MetaLearningFL} and relation classification~\cite{Han2018FewRelAL}. The closest related works to ours is~\cite{Fritzler2019FewshotCI} which explores prototypical network on few-shot NER, but only utilizes RNNs as the backbone model and does not leverage the power of large-scale Transformer-based architectures for word representations. 
% Moreover, they do not test their methods on out-of-domain entities, as the entities used for training and testing are from the same dataset.
Our work is similar to~\cite{Ziyadi2020ExampleBasedNE,Wiseman2019LabelAgnosticSL} in that all of them utilize the nearest neighbor criterion to assign the entity type, but differs in that ~\cite{Ziyadi2020ExampleBasedNE,Wiseman2019LabelAgnosticSL} consider every individual token instance for nearest neighbor comparison, while ours considers prototypes for comparison. Hence, our method is much more scalable when the number of given examples increases.
% also explored some simplified variants of prototype-based methods for NER. Both methods are designed for zero-shot NER, and use instance-wise nearest neighbor classification methods leveraging all available training data, thus are not scalable for fine-tuning models in few-shot learning setting. 
% Our work is different in two aspects: (1) The
% \CL{describe the differences}

\paragraph{Supervised pre-training.} In computer vision, it is a {\em de facto} standard to transfer ImageNet-supervised pre-trained models to small image datasets to pursue high recognition accuracy ~\cite{yosinski2014transferable}. The recent work named big transfer~\cite{kolesnikov2019big} has achieved SoTA on various vision tasks via pre-training on billions of noisily labeled web images. To gain a stronger transfer learning ability, one may combine supervised and self-supervised methods~\cite{li2020mopro,li2020hexa}. In NLP, supervised/grounded pre-training have been recently explored for natural language generation (NLG)~\cite{keskar2019ctrl,zellers2019defending,peng2020few,gao2020robust,li2020optimus}. They aim to endow GPT-2~\cite{gpt2}, an ability of enabling high-level semantic controlling in language generation, and are often pre-trained on massive corpus consisting of text sequences associated with prescribed codes such as text style, content description, and task-specific behavior. 
% They can viewed as the inverse process of supervised pre-training of NLU.
In contrast to NLG, to our best knowledge, large-scale supervised pre-training has been little studied for natural language understanding (NLU). There are early works showing promising results by transferring from medium-sized datasets to small datasets in some NLU applications; For example, from MNLI to RTE for sentence classification~\cite{phang2018sentence,clark2020electra,an2020repulsive}, and from OntoNER to CoNLL for NER~\cite{Yang2020SimpleAE}. Our work further increases the supervised pre-training at the scale of web data~\cite{Ghaddar2017WiNERAW}, 1000 orders of magnitude larger than~\cite{Yang2020SimpleAE}, showing consistent improvements.

\paragraph{Self-training.}
Self-training~\cite{scudder1965probability} is one of the earliest semi-supervised methods, and has recently achieved improved performance for tasks such as ImageNet classification~\cite{Xie2020SelfTrainingWN}, visual object detection~\cite{zoph2020rethinking}, neural machine translation~\cite{he2019revisiting}, 
sentence classification~\cite{mukherjee2020uncertainty,du2020self}. It is shown via object detection tasks in~\cite{zoph2020rethinking} that stronger data augmentation and more labeled data can diminish the value of pre-training, while self-training is always helpful in both low-data and high-data regimes. Our work presents the first study of self-training for NER, and we observe similar phenomenons: it {\em consistently} boosts few-shot learning performance across all 10 datasets.

\section{Experiments}
In this section, we first compare the performance of different combinations of the three schemes on 10 benchmark datasets with various proportions of training data, and then compare our approaches with SoTA methods proposed for settings of few-shot learning and immediate inference for unseen entity types.

\begin{table*}[t!]
\centering
\small
\scalebox{0.86}{
\begin{tabular}{c|c|c|c|c|c|c|c|c|c|c}
\toprule
Datasets & CoNLL & Onto & WikiGold & WNUT & Movie & Restaurant & SNIPS & ATIS & Multiwoz & I2B2\\ 
\midrule
Domain & News & General & General & Social Media & Review & Review & Dialogue & Dialogue & Dialogue & Medical \\
$\#$Train & 14.0k & 60.0k & 1.0k & 3.4k & 7.8k & 7.7k & 13.6k & 5.0k & 20.3k & 56.2k \\ 
$\#$Test & 3.5k & 8.3k & 339 & 1.3k & 2.0k & 1.5k & 697 & 893 & 2.8k & 51.7k \\ 
$\#$Entity Types & 4 & 18 & 4 & 6 & 12 & 8 & 53 & 79 & 14 & 23 \\
\bottomrule
\end{tabular}}
\vspace{-2mm}
\caption{Statistics on the 10 public datasets studied in our NER benchmark.}
\label{tab:stats}
\vspace{-2mm}
\end{table*}

\subsection{Settings}
\paragraph{Methods.}
Throughout our experiments, the pre-trained base RoBERTa model is employed as the backbone network. We investigate the following 6 schemes for the comparative study:
$(\RN{1})$ \textbf{LC} is the {\em linear classifier} fine-tuning method in Section 2, \ie adding a linear classifier on the backbone, and directly fine-tuning on entire model on the target dataset; 
% \jg{``native fine-tuning'' is a understatement. how about ``standard or classical'' fine-tuning?}
$(\RN{2})$ \textbf{P}  indicates the {\em prototype-based method} in Section 3.1;  
$(\RN{3})$ \textbf{NSP} refers to the {\em noisy supervised pre-training} in Section 3.2; Depending on the pre-training objective, we have \textbf{LC+NSP} and \textbf{P+NSP}. 
$(\RN{4})$ \textbf{ST} is the {\em self-training} approach in Section 3.3, it is combined with {\em linear classifier} fine-tuning, denoted as \textbf{LC+ST};
$(\RN{5})$ \textbf{LC+NSP+ST}.

\paragraph{Datasets.}
We evaluate our methods on 10 public benchmark datasets, covering a wide range of domains: OntoNotes 5.0~\cite{Ontonotes},  WikiGold\footnote{ \url{https://github.com/juand-r/entity-recognition-datasets}}~\cite{Balasuriya2009NamedER} on general domain, 
CoNLL 2003~\cite{Sang2003IntroductionTT} on news domain,
WNUT 2017~\cite{Derczynski2017ResultsOT} on social domain,
MIT Moive~\cite{Liu2013QueryUE} and MIT Restaurant\footnote{ \url{https://groups.csail.mit.edu/sls/downloads/}}~\cite{Liu2013AsgardAP} on review domain,
SNIPS\footnote{\url{https://github.com/snipsco/nlu-benchmark/tree/master/2017-06-custom-intent-engines}}~\cite{Coucke2018SnipsVP}, ATIS\footnote{\url{ https://github.com/yvchen/JointSLU}}~\cite{HakkaniTr2016MultiDomainJS} and Multiwoz\footnote{\url{https://github.com/budzianowski/multiwoz}}~\cite{budzianowski2018large} on dialogue domain,
and I2B2\footnote{\url{https://portal.dbmi.hms.harvard.edu/projects/n2c2-2014/}}~\cite{Stubbs2015AnnotatingLC} on medical domain. 
The detailed statistics of these datasets are summarized in Table \ref{tab:stats}. 

For each dataset, we conduct three sets of experiments using various proportions of the training data: 5-shot, 10\% and 100\%. For 5-shot setting, we sample 5 sentences for each entity type in the training set and repeat each experiment for 10 times. For 10\% setting, we down-sample 10 percent of the training set, and for 100\% setting, we use the full training set as labeled data. We only study the self-training method in 5-shot and 10\% settings, by using the rest of the training set as unlabeled in-domain corpus.

\paragraph{Hyper-parameters.}
We have described details for noisy supervised pre-training in Section \ref{sec:nsp}. For training on target datasets, we set a fixed set of hyperparameters across all the datasets: For the linear classifier, we set batch size = $16$ for 100\% and 10\% settings, batch size = $4$ for 5-shot setting. For each episode in the prototype-based method, we set the number of sentences per entity type in support and query set $(K, K')$ to be $(5, 15)$ for 100\% and 10\% settings, and $(2, 3)$ for 5-shot setting.
For both training objectives, we set learning rate = $5e^{-5}$ for 100\% and 10\% settings, and learning rate = $1e^{-4}$ for 5-shot setting.
For all training data sizes, we set training epoch = $10$, and Adam optimizer~\cite{Kingma2015AdamAM} is used with the same linear decaying schedule as the pre-training stage. For self-training, we set $\lambda_{\mathtt{U}}=0.5.$

\paragraph{Evaluation.}

We follow the standard protocols for NER tasks to evaluate the performance on the test set~\cite{Sang2003IntroductionTT}. Since RoBERTa tokenizes each word into subwords, we generate word-level predictions based on the first word piece of a word. Word-level predictions are then turned into entity-level predictions for evaluation when calculating the f1-score.
Two tagging schemas are typically considered to encode chunks of tokens into entities: 
BIO schema marks the beginning token of an entity as \texttt{B-X} and the consecutive tokens as \texttt{I-X}, and other tokens are marked as \texttt{O}. IO schema uses \texttt{I-X} to mark all tokens inside an entity, thus is more defective as there is no boundary tag. In our study, we  use BIO schema by default, but report the performance evaluated by IO schema for fair comparison with some previous studies.
% When doing inference in prototype-based methods, class prototypes are calculated by the average of all available training data in each training setting.

\begin{table*}[t!]
    \centering
    \small
    \begin{tabular}{c|c|C{1.5cm}|C{1.5cm}|C{1.5cm}|C{1.5cm}|C{1.5cm}|C{2.0cm}}
    \toprule
     \multirow{2}{*}{Datasets}  & \multirow{2}{*}{Settings}  & \circled{1} & \circled{2} & \circled{3} & \circled{4} & \circled{5} & \circled{6}  \\ [3pt]
     &  & \hlightPyellow{\text{LC}} & \hlightPyellow{\text{LC}}~~+~~~\hlightPblue{\text{NSP}} & \hlightPred{\text{P}} & \hlightPred{\text{P}}~ +~~\hlightPblue{\text{NSP}} & \hlightPyellow{\text{LC}}~~+~~\hlightPgreen{\text{ST}} & \hlightPyellow{\text{LC}}~~+~~\hlightPblue{\text{NSP}}~~+~~\hlightPgreen{\text{ST}}  \\
    \midrule
    \multirow{3}{*}{CoNLL} & 5-shot & 0.535 & 0.614  & 0.584 & 0.609%/0.681
    & 0.567  & \textbf{0.654} \\
    & 10\% & 0.855 & 0.891 & 0.878 & 0.888 & 0.878 & \textbf{0.895} \\
    & 100\% & 0.919 & \textbf{0.920} & 0.911 & 0.915 & - & - \\
    \midrule
    \multirow{3}{*}{Onto} & 5-shot & 0.577 & 0.688 & 0.533 & 0.570%/0.465 
    & 0.605 & \textbf{0.711} \\
    & 10\% & 0.861 & \textbf{0.869} & 0.854 & 0.846 & 0.867 & 0.867 \\
    & 100\% & 0.892 & \textbf{0.899} & 0.886 & 0.883 & - & - \\
    \midrule
    \multirow{3}{*}{WikiGold} & 5-shot & 0.470 & 0.640 & 0.511 & 0.604%/0.691 
    & 0.481 & \textbf{0.684} \\
    & 10\% & 0.665 & 0.747 & 0.692 & 0.701 & 0.695 & \textbf{0.759} \\
    & 100\% & 0.807 & \textbf{0.839} & 0.801 & 0.827 & - & - \\
    \midrule
    \multirow{3}{*}{WNUT17} & 5-shot & 0.257 & 0.342 & 0.295 & 0.359%/0.347 
    & 0.300  & \textbf{0.376} \\
    & 10\% & 0.483 & 0.492 & 0.485 & 0.478 & 0.490 & \textbf{0.505} \\
    & 100\% & 0.489 & 0.520 & 0.552 & \textbf{0.560} & - & - \\
    \midrule
    \multirow{3}{*}{MIT Movie} & 5-shot & 0.513 & 0.531 & 0.380 & 0.438%/0.201 
    & 0.541 & \textbf{0.559} \\
    & 10\% & 0.651 & 0.657 & 0.563 & 0.583 & 0.659 & \textbf{0.666} \\
    & 100\% & \textbf{0.693} & 0.692 & 0.632 & 0.641 & - & - \\
    \midrule
    \multirow{3}{*}{MIT Restaurant} & 5-shot & 0.487 & 0.491  & 0.441 & 0.484%/0.201  
    & 0.503 & \textbf{0.513} \\
    & 10\% & 0.745 & 0.734 & 0.713 & 0.721 & \textbf{0.750} & 0.741 \\
    & 100\% & 0.790 & \textbf{0.793} & 0.787 & 0.791 & - & - \\
    \midrule
    \multirow{3}{*}{SNIPS} & 5-shot & 0.792  & 0.824  & 0.750  & 0.773%/0.581  
    & 0.796 & \textbf{0.830} \\
    & 10\% & 0.945 & \textbf{0.950} & 0.879 & 0.896  & 0.946 & 0.942 \\
    & 100\% & 0.970 & \textbf{0.972} & 0.923 & 0.956 & - & - \\
    \midrule
    \multirow{3}{*}{ATIS} & 5-shot & \textbf{0.908}  & \textbf{0.908} & 0.842 & 0.896%/0.732  
    & 0.904 & 0.905 \\
    & 10\% & 0.883 & 0.898 & 0.785 & 0.896 & 0.898 & 
    \textbf{0.903} \\
    & 100\% & 0.953 & \textbf{0.956} & 0.929 & 0.943 & - & - \\
    \midrule
    \multirow{3}{*}{Multiwoz} & 5-shot & 0.123  & 0.198 & 0.219  & \textbf{0.451}%/0.201  
    & 0.200  & 0.225 \\
    & 10\% & 0.826 & 0.830 & 0.787 & 0.805 & 0.835 & \textbf{0.841} \\
    & 100\% & 0.880 & \textbf{0.885} &  0.837 & 0.845 & - & - \\
    \midrule
    \multirow{3}{*}{I2B2} & 5-shot & 0.360 \iffalse0.068\fi & 0.385\iffalse0.034\fi & 0.320\iffalse0.064\fi  & 0.366\iffalse0.047\fi %/0.201
    &  0.365 \iffalse0.058\fi & \textbf{0.393}\iffalse0.029\fi \\
    & 10\% & 0.855 & 0.869 & 0.703 & 0.762 & 0.865 & \textbf{0.871} \\
    & 100\% & 0.932 & \textbf{0.935} &  0.895 & 0.906 & - & - \\
    \midrule
    \multirow{3}{*}{{\bf Average}} & 5-shot & 0.502  &0.562 &  0.488 &  0.555  & 0.526   & \textbf{0.585} \\
    & 10\% & 0.777 & 0.794 & 0.734 & 0.758 & 0.788 & \textbf{0.799} \\
    & 100\% & 0.833 & \textbf{0.841} &  0.815 & 0.827 & - & - \\
    \midrule
    \end{tabular}
    \vspace{-2mm}
    \caption{F1-score on benchmark datasets with various sizes of training data. \hlightPyellow{\text{LC}}~ is {\it linear classifier} fine-tuning method, ~~\hlightPred{\text{P}}~ is {\it prototype-based training} using a nearest neighbor objective, ~~\hlightPblue{\text{NSP}}~ is {\it noising supervised pre-training} and ~~\hlightPgreen{\text{ST}}~ is {\it self-training}. The best results are in {\bf bold}.}
    \label{tab:main_results}
    \vspace{-2mm}
\end{table*}

\subsection{Comprehensive Comparison Results}
To gain thorough insights and benchmark few-shot NER, we first perform an extensive comparative study on 6 methods across 10 datasets. The results are shown in Table \ref{tab:main_results}.
% shows the F1-score of 6 combinations of training schemes on 9 datasets, each including 3 sizes of training data. 
We can draw the following major conclusions: 
$(\RN{1})$
By comparing column {\small\circled{1}} and {\small\circled{2}} (or comparing {\small\circled{3}} and {\small\circled{4}}), it clearly shows that noisy supervised pre-training provides better results in most datasets, especially in the 5-shot setting, which demonstrates that \textbf{NSP} endows the model an ability to extract better NER-related features. 
$(\RN{2})$
The comparison between column {\small\circled{1}} and {\small\circled{3}} provides a head-to-head comparison between linear classifier and prototype-based methods: while the prototype-based method demonstrates better performance than \textbf{LC} on CoNLL, WikiGold, WNUT17 and Multiwoz in the 5-shot learning setting, it falls behind \textbf{LC} on other datasets and in average statistics. It shows that the prototype-based method only yields better results when there is very limited labeled data: the size of both entity types and examples are small. 
$(\RN{3})$
When comparing column {\small\circled{5}} with {\small\circled{1}}  (or comparing column {\small\circled{6}} and {\small\circled{2}}), we observe that using self-training consistently works better than directly fine-tuning with labeled data only, suggesting that \textbf{ST} is a useful technique to leverage in-domain unlabeled data if allowed. 
$(\RN{4})$ Column {\small\circled{6}} shows the highest F1-score in most cases, demonstrating the three proposed schemes in this paper are complementary %/orthogonal 
to each other, and can be combined to yield best results in practice.

\begin{table*} [t!]
\centering
\small
%\addtolength{\tabcolsep}{-2pt}
\begin{tabular}{ll|lll|l}
    \toprule
   Schema &  Methods & CoNLL & I2B2 & WNUT & Average \\ \midrule
   \multirow{6}{*}{IO}
     & SimBERT $^\dagger$  & 0.286\tiny{${\pm}$0.025} &  0.091\tiny{${\pm}$0.007} & 0.077\tiny{${\pm}$0.022} & 0.151 \\
     & L-TapNet+CDT $^\dagger$  & 0.671\tiny{${\pm}$0.016} &  0.101\tiny{${\pm}$0.009} & 0.238\tiny{${\pm}$0.039} & 0.336 \\
     & StructShot $^\dagger$  & 0.752\tiny{${\pm}$0.023} & 0.318\tiny{${\pm}$0.018} & 0.272\tiny{${\pm}$0.067} & 0.447  \\
     & \text{P}~+~\text{NSP}   & 0.757\tiny{${\pm}$0.021} & 0.322\tiny{${\pm}$0.033} & 0.442\tiny{${\pm}$0.024} & 0.507 \\
     & \text{LC}~+~\text{NSP}  & 0.771\tiny{${\pm}$0.035} & 0.371\tiny{${\pm}$0.035} & 0.417\tiny{${\pm}$0.022} & \textbf{0.520} \\
     & \text{LC}~+~\text{NSP}~+~\text{ST}   & 0.779\tiny{${\pm}$0.040} & 0.376\tiny{${\pm}$0.028} & 0.419\tiny{${\pm}$0.028} & \textbf{0.525} \\     
     \midrule
      \multirow{3}{*}{BIO}
     & \text{P}~+~\text{NSP}   & 0.756\tiny{${\pm}$0.017} & 0.334\tiny{${\pm}$0.024} & 0.424\tiny{${\pm}$0.012} & 0.505 \\
     & \text{LC}~+~\text{NSP}  & 0.712\tiny{${\pm}$0.048} & 0.364\tiny{${\pm}$0.032} & 0.403\tiny{${\pm}$0.029} & 0.493 \\     
     & \text{LC}~+~\text{NSP}~+~\text{ST} & 0.722\tiny{${\pm}$0.011} & 0.369\tiny{${\pm}$0.021} & 0.409\tiny{${\pm}$0.013} & 0.500 \\
    \bottomrule
\end{tabular}
\vspace{-0mm}
\caption{Comparison of F1-score with SoTA on 5-shot NER tasks. Results of both BIO and IO schemas are reported for fair comparison. The best results are in {\bf bold}. $^\dagger$ indicates results from~\cite{Yang2020SimpleAE}.}
\label{tab:5shot-results}
\vspace{2mm}
\end{table*}

\begin{table*}[h]
\centering
\small
\scalebox{0.92}{
\begin{tabular}{c|c|c|c|c|c|c|c}
\toprule
\multirow{2}{*}{Datasets}&\multirow{2}{*}{Methods}&\multicolumn{6}{c}{\textit{Number of support examples per entity type}}\\\cline{3-8}
&&\textit{10}&\textit{20}&\textit{50}&\textit{100}&\textit{200}&\textit{500}\\
\midrule

\multirow{6}{*}{ATIS}&Neigh.Tag.$^\dagger$ & 0.067\tiny{$\pm$0.008} & 0.088\tiny{$\pm$0.007} & 0.111\tiny{$\pm$0.007}& 0.143\tiny{$\pm$0.006}& 0.221\tiny{$\pm$0.006}& 0.339\tiny{$\pm$0.006}\\
&Example$^\dagger$ & 0.174\tiny{$\pm$0.011} & 0.198\tiny{$\pm$0.012} & 0.222\tiny{$\pm$0.011} & 0.268\tiny{$\pm$0.027} & 0.345\tiny{$\pm$0.022} & 0.401\tiny{$\pm$0.010}\\
&Prototype & 0.381\tiny{$\pm$0.021} & 0.391\tiny{$\pm$0.022} & 0.376\tiny{$\pm$0.008} & 0.379\tiny{$\pm$0.005} & 0.377\tiny{$\pm$0.006} & 0.376\tiny{$\pm$0.003} \\
&Prototype + NSP & 0.684\tiny{$\pm$0.013} & 0.712\tiny{$\pm$0.014} & 0.716\tiny{$\pm$0.013} & 0.705\tiny{$\pm$0.010} & 0.705\tiny{$\pm$0.006} & 0.708\tiny{$\pm$0.002}\\
&Multi-Prototype & 0.339\tiny{$\pm$0.016} & 0.362\tiny{$\pm$0.018} & 0.366\tiny{$\pm$0.005} & 0.373\tiny{$\pm$0.004} & 0.371\tiny{$\pm$0.005} & 0.372\tiny{$\pm$0.003}\\
&Multi-Prototype + NSP & \textbf{0.712}\tiny{$\pm$0.014} & \textbf{0.748}\tiny{$\pm$0.011} & \textbf{0.760}\tiny{$\pm$0.008} & \textbf{0.742}\tiny{$\pm$0.005} & \textbf{0.743}\tiny{$\pm$0.003} & \textbf{0.746}\tiny{$\pm$0.002}\\
\hline

\multirow{6}{*}{MIT.Restaurant}&Neigh.Tag.$^\dagger$& 0.042\tiny{$\pm$0.018} & 0.038\tiny{$\pm$0.008} & 0.037\tiny{$\pm$0.007} & 0.046\tiny{$\pm$0.008} & 0.055\tiny{$\pm$0.011} & 0.081\tiny{$\pm$0.006}\\
&Example.$^\dagger$ & 0.276\tiny{$\pm$0.018} & 0.295\tiny{$\pm$0.010} & 0.312\tiny{$\pm$0.007} & 0.337\tiny{$\pm$0.005} & 0.345\tiny{$\pm$0.004} & 0.346 \\
&Prototype & 0.330\tiny{$\pm$0.013} & 0.332\tiny{$\pm$0.013} & 0.332\tiny{$\pm$0.010} & 0.329\tiny{$\pm$0.003} & 0.329\tiny{$\pm$0.004} & 0.331\tiny{$\pm$0.003}\\
&Prototype + NSP & 0.455\tiny{$\pm$0.016} & 0.455\tiny{$\pm$0.012} & 0.455\tiny{$\pm$0.013} & 0.438\tiny{$\pm$0.013} & 0.437\tiny{$\pm$0.008} & 0.438\tiny{$\pm$0.006}\\
&Multi-Prototype & 0.345\tiny{$\pm$0.012} & 0.360\tiny{$\pm$0.015} & 0.371\tiny{$\pm$0.012} & 0.376\tiny{$\pm$0.009} & 0.385\tiny{$\pm$0.005} & 0.386\tiny{$\pm$0.004}\\
&Multi-Prototype + NSP & \textbf{0.461}\tiny{$\pm$0.019} & \textbf{0.482}\tiny{$\pm$0.011} & \textbf{0.496}\tiny{$\pm$0.008} & \textbf{0.496}\tiny{$\pm$0.011} & \textbf{0.500}\tiny{$\pm$0.005} & \textbf{0.501}\tiny{$\pm$0.003}\\
\hline
\multirow{6}{*}{MIT Movie}&Neigh.Tag.$^\dagger$& 0.031\tiny{$\pm$0.020} & 0.045\tiny{$\pm$0.019} & 0.041\tiny{$\pm$0.011} & 0.053\tiny{$\pm$0.009} & 0.054\tiny{$\pm$0.007} & 0.086\tiny{$\pm$0.008} \\
&Example.$^\dagger$ & \textbf{0.401}\tiny{$\pm$0.011} & \textbf{0.395}\tiny{$\pm$0.007} & \textbf{0.402}\tiny{$\pm$0.007} & \textbf{0.400}\tiny{$\pm$0.004} & \textbf{0.400}\tiny{$\pm$0.005} & \textbf{0.395}\tiny{$\pm$0.007} \\
&Prototype & 0.175\tiny{$\pm$0.007} & 0.168\tiny{$\pm$0.006} & 0.170\tiny{$\pm$0.004} & 0.174\tiny{$\pm$0.003} & 0.173\tiny{$\pm$0.002} & 0.173\tiny{$\pm$0.002}\\
&Prototype + NSP & 0.303\tiny{$\pm$0.011} & 0.293\tiny{$\pm$0.007} & 0.285\tiny{$\pm$0.006} & 0.284\tiny{$\pm$0.002} & 0.282\tiny{$\pm$0.002} & 0.280\tiny{$\pm$0.002}\\
&Multi-Prototype & 0.197\tiny{$\pm$0.007} & 0.207\tiny{$\pm$0.005} & 0.219\tiny{$\pm$0.004} & 0.227\tiny{$\pm$0.002} & 0.229\tiny{$\pm$0.003} & 0.230\tiny{$\pm$0.002}\\
&Multi-Prototype + NSP & 0.364\tiny{$\pm$0.020} & 0.368\tiny{$\pm$0.011} & 0.380\tiny{$\pm$0.006} & 0.382\tiny{$\pm$0.003} & 0.354\tiny{$\pm$0.003} & 0.383\tiny{$\pm$0.002}\\
\bottomrule

% \multirow{2}{*}{Mixed Domain}&Neigh.Tag.& 0.045\tiny{$\pm$0.008} & 0.055\tiny{$\pm$0.005} & 0.067\tiny{$\pm$0.006} & 0.087\tiny{$\pm$0.004} & 0.131\tiny{$\pm$0.006} & 0.203\tiny{$\pm$0.004}\\
% &Example. & 0.166\tiny{$\pm$0.011} & 0.203\tiny{$\pm$0.008} & 0.235\tiny{$\pm$0.006} & 0.274\tiny{$\pm$0.012} & 0.322\tiny{$\pm$0.012} & 0.359\tiny{$\pm$0.006} \\
% \hline
\end{tabular}}
\caption{F1-score on training-free settings, \ie predicting novel entity types using nearest neighbor methods. The best results are in {\bf bold}. $^\dagger$ indicates results from ~\citep{Ziyadi2020ExampleBasedNE,Wiseman2019LabelAgnosticSL}. }
\label{tab:zero_shot}
\end{table*}

\subsection{Comparison with SoTA Methods}
\paragraph{Competitive methods.} The current SoTA on few-shot NER includes:
$(\RN{1})$
{\it StructShot}~\cite{Yang2020SimpleAE}, which extends the nearest neighbor classification with a decoding process using abstract tag transition distribution. Both the model and the transition distribution are trained from the source dataset OntoNotes. 
$(\RN{2})$
{\it L-TapNet+CDT}~\cite{Hou2020FewshotST} is a slot tagging method which constructs an embedding projection space using label name semantics to well separate different classes. It also includes a collapsed dependency transfer mechanism to transfer label
dependency information from source domains to
target domains.
$(\RN{3})$
{\it SimBERT}~is a simple baseline reported in~\cite{Yang2020SimpleAE,Hou2020FewshotST}; it utilizes a nearest neighbor classifier based on the contextualized representation output by the pre-trained BERT, without fine-tuning on few-shot examples. The results reported in the StructShot paper use IO schema instead of BIO schema, thus we report our performance on both for completeness. 
% Besides CoNLL and WNUT, a new medical dataset I2B2 is used in the experiments \cite{Yang2020SimpleAE}, which contains 23 entity types along with 56.2k training sentences and 51.7k testing sentences.  We follow an existing preprocessing steps\footnote{\url{https://github.com/Franck-Dernoncourt/NeuroNER/tree/master/neuroner/data/i2b2_2014_deid}}. \CL{I am wondering if we could add I2B2 in Table 1}

% \paragraph{Results}

For fair comparison, following \cite{Yang2020SimpleAE}, we also continuously pre-train our model on OntoNotes after the noisy supervised pre-training stage.
For each 5-shot learning task, we repeat the experiments 10 times by re-sampling few-shot examples each time. The results are reported in Table \ref{tab:5shot-results}. We observe that our proposed methods consistently outperform the StructShot model across all three datasets, even by simply pre-training the model on large-scale noisily tagged datasets like Wikipedia. 
Our best model outperforms the previous SoTA by 8\% F1-score, which demonstrates that using large amounts of unlabeled in-domain corpus is promising for enhancing the few-shot NER performance.

\subsection{Training-free Method Comparison}
% \jg{It is controversial to call the setting ``zero-shot'' because we do use in-domain labels to build models no matter these models are parametric (e.g., linear classifier) or non-parametric (nearest-neighbor). Note that the ``setting'' is independent of the models being developed. As long as the model \emph{sees} the label (no matter how these labels are used), this is NOT a zero-shot learning. For example, in the GPT-3 paper, the same setting is called ``few-shot learning'' where the model sees a few examples of the task without model parameter update, as in Figure 2.1. Instead, ``zero-shot'' learning does not allow any demonstrations. }
% 
Some real-world applications require immediate inference on unseen entity types. For example, novel entity types with a few examples are frequently given in an online fashion, but updating model weights $\thetav$ frequently is prohibitive. One may store some token examples as supports and utilize them for nearest neighbor classification. The setting is referred to as {\it training-free} in~\cite{Wiseman2019LabelAgnosticSL,Ziyadi2020ExampleBasedNE}, as the models identify new entities in a
completely unseen target domain using only a few
supporting examples in this new domain, without updating $\thetav$ in that target domain. Our prototype-based method is able to perform such immediate inference. Two recent works on training-free NER are: 
$(\RN{1})$
{\it Neighbor-tagging}~\cite{Wiseman2019LabelAgnosticSL} copies token-level labels from weighted nearest neighbors;
$(\RN{2})$
{\it Example-based NER}~\cite{Ziyadi2020ExampleBasedNE} is the SoTA on training-free NER, which identifies the starting and ending tokens of unseen entity types.

We observed that our basic prototype-based method, under the training-free setting, does not gain from more given examples. We hypothesize that this is because tokens belonging to the same entity type are not necessarily close to each other, and are often separated in the representation space. 
Though it is hard to find one single centroid for all tokens in the same type, we assume that there exist local clusters of tokens belonging to the same type.
To resolve such issue, we follow~\cite{deng2020sub} and extend our method to a version called {\em Multi-Prototype}, by creating $K/5$ prototypes for each type given $K$ examples per type. (\eg 2 prototypes per class are used for the 10-shot setting). 
% Specifically, input examples are randomly divided into $V$ 5-shot groups, and prototype for class $m$ in group $j$ is notated as $\cv_m^j$. Suppose there are 10 examples for each class, then $V=10/5=2$. 
The prediction score for a testing token belonging to a type is computed via averaging the prediction probabilities from all prototypes of the same type.
%
% \begin{equation}
% {\small 
%     q_{\text{multi}}(\yv\!=\!\I_m|\xv)\!=\!\frac{\sum_{j=1}^V \text{SoftMax}(-d(f_{\thetav_0}(\xv), \cv_m^j))}{V}
% }
% \end{equation}
%

We compare with previous methods in Table \ref{tab:zero_shot}. 
% We also report the results of using multi-prototype methods with or without noisy supervised pre-training in Table \ref{tab:zero_shot}. 
We observe that multi-prototype methods not only benefit from more support examples, but also surpass neighbor tagging methods and example-based NER by a large margin on two out of three datasets. For the MIT Movie dataset, one entity type can span a large chunk with multiple consecutive words in a sentence, which favors the span-based method like~\cite{Ziyadi2020ExampleBasedNE}. For example, the underlined part in the sentence ``\textit{what movie does the quote \dotuline{i dont think we are in kansas anymore} come from}'' is annotated as entity type \texttt{Quote}. The proposed methods in this paper can be combined with the span-based approach to specifically tackle this problem, and we leave it as future work. Further, if slightly fine-tuning is allowed, we see that the prototype-based method achieves 0.438 with 5-shot learning in Table \ref{tab:main_results}, better than 0.395 achieved by example-based NER given 500 examples.

% though our proposed methods do not work better than the example-based NER in the zero-shot setting, we show in Table \ref{tab:main_results} that by a little fine-tuning on only 5-shot examples using the prototype-based method, we can already achieve 0.438, better than 0.395 achieved by example-based NER given 500 examples.

% \CL{I don't understand the last sentence} I want to point out that in table 2, the column \circled{4} of row MIT Movie (5-shot) is 0.438, which only requires little fine-tuning, but performs much better than example-based ner with 500 examples (0.395). 

\section{Conclusion}
We have presented a comprehensive study on few-shot NER. Three foundational methods and their combinations are systematically investigated: prototype-based methods, noisy supervised pre-training and self-training. They are intensively compared on 10 public datasets under various settings. All of them can improve the PLM's generalization ability when learning from a few labeled examples, among which supervised pre-training and self-training turn out to be particularly effective. The proposed schemes achieve SoTA on both few-shot and training-free settings compared with recently proposed methods. We will release our benchmarks and code for few-shot NER, and hope that it can inspire future research with more advanced methods to tackle this challenging and practical problem.

\bibliographystyle{acl_natbib}
\bibliography{acl2020}
\end{document}